\documentclass{article}

\usepackage[final]{neurips_2025}
\usepackage{amssymb}
\usepackage{makecell}

\usepackage{subfigure}
\usepackage{multirow}
\usepackage{graphicx}
\usepackage[utf8]{inputenc} 
\usepackage[T1]{fontenc}    
\usepackage{hyperref}       
\usepackage{url}            
\usepackage{booktabs}       
\usepackage{amsfonts}       
\usepackage{nicefrac}       
\usepackage{microtype}      
\usepackage{xcolor}         
\usepackage{array} 
\usepackage[most]{tcolorbox}

\title{Fin-o1: On the Transferability of Reasoning-Enhanced LLMs and Reinforcement Learning to Finance}

\author{%
  Lingfei Qian \\
  The FinAI\\
  \texttt{lfqian94@gmail.com} 
  \And
  Yan Wang \\
  The FinAI 
  \And
  Xueqing Peng \\
  The FinAI 
  \And
  Weipeng Zhou \\
  The FinAI 
  \And
  Yi Han \\
  Georgia Institute of Technology 
  \And
  Yilun Zhao \\
  Yale University
  \And
  Jimin Huang \\
  The FinAI
  \And  
  Qianqian Xie\thanks{Corresponding author} \\
  The FinAI\\
  \texttt{xqq.sincere@gamil.com}
  \And
  Jian-Yun Nie\\
  University of Montreal
}

\begin{document}

\maketitle

\begin{abstract}
As the fundamental capability behind decision-making in finance, financial reasoning poses distinct challenges for LLMs.
Although reinforcement learning (RL) have boosted generic reasoning, the progress in finance is hindered by the absence of empirical study of building effective financial chain-of-thought (CoT) corpus, a systematic comparison of different RL methods, and comprehensive benchmarks. 
To address these gaps, we introduce FinCoT, the first open high-fidelity CoT corpus for finance, distilled from seven QA datasets by a novel three-stage pipeline that incorporates domain supervision, iterative LLM refinement, and difficulty-aware filtering. 
Based on FinCoT, we develop Fin-o1, the first open financial reasoning models trained via supervised fine-tuning and GRPO-based RL.
Our models outperform existing financial reasoning models and SOTA general models such as GPT-o1, DeepSeek-R1, and GPT-4.5. 
We also investigate the effectiveness of three different RL methods in improving domain-specific reasoning, offering the first such empirical study.
We finally propose FinReason, the first financial reasoning benchmark covering multi-table analysis, long-context reasoning, and equation-based tasks, and evaluate 29 LLMs.
Our extensive experiments reveal general reasoning models excel on standard benchmarks yet exhibit obvious performance degradation in financial contexts; even finance-tuned models like Dianjin-R1 and FinR1 degrade on lengthy documents. In contrast, our Fin-o1 models consistently outperform their backbones and larger GPT-o1 and DeepSeek-R1, confirming the effectiveness of our data building and model training strategy. Our study further shows that GRPO yields reliable gains whereas PPO and DPO do not, highlighting the need for targeted data and optimisation rather than scale alone.
These findings show that financial reasoning has unique challenges unmet by mere model scaling or generic techniques, highlighting the need for specialized data, models, and optimization.
We release all datasets, models, and codes to support future research.~\footnote{
Dataset: \url{https://huggingface.co/datasets/TheFinAI/FinCoT} \\
\hspace*{1.5em} Fin-o1: \url{https://huggingface.co/TheFinAI/Fin-o1-8B} \\
\hspace*{4.7em}         \url{https://huggingface.co/TheFinAI/Fin-o1-14B} \\
\hspace*{1.5em} Code: \url{https://github.com/The-FinAI/Fino1}
}
\end{abstract}

\section{Introduction} 
\label{intro}
Financial reasoning is a critical yet underexplored capability for large language models (LLMs), especially as they are increasingly deployed in high-stakes financial applications. Unlike general reasoning tasks, such as mathematical problem solving or code generation, financial reasoning requires models to interpret domain-specific language, analyze structured data like financial tables~\cite{xie2024open}, and extract insights from lengthy, complex documents~\cite{reddy2024docfinqa}. It demands not only numerical computation, but also contextual understanding of financial regulations, economic concepts, and ambiguous textual information~\cite{xie2023pixiu, xie2024finben}.

Recent advances in reinforcement learning (RL) methods,
including proximal policy optimization 
(PPO)~\cite{schulman2017proximal}, direct preference optimization (DPO)~\cite{rafailov2023direct}, and generalized reinforcement preference optimization (GRPO)~\cite{tang2024generalized},
have substantially enhanced the reasoning capabilities of large language models. These RL-based approaches form the foundation of state-of-the-art (SOTA) reasoning models such as OpenAI’s o1~\cite{jaech2024openai} and DeepSeek’s R1~\cite{guo2025deepseek}, achieving notable improvements on mathematical and logical problem-solving benchmarks~\cite{temsah2024openai, zhong2024evaluation}. By directly optimizing for reasoning performance, RL strategies have become essential for advancing the complex reasoning abilities of LLMs.

However, improving LLMs with RL for financial reasoning is yet hindered by three critical challenges (As shown in Table \ref{combined_table}). First, the absence of high-quality, publicly available chain-of-thought (CoT) datasets for financial reasoning limits the ability to model domain-specific reasoning processes. Existing resources are either proprietary~\cite{jaech2024openai} or adapted from general domains~\cite{guo2025deepseek, yang2024qwen2}, failing to capture the nuances of financial analysis.
No empirical study has asked which source documents or path-generation strategies actually produce effective financial CoT traces.
Second, while RL methods have enhanced general reasoning, their effectiveness in financial tasks remains unexplored, with no open-source models demonstrating domain-adapted improvements. 
The impact of different RL algorithms on financial reasoning remains untested: RL methods like PPO, DPO and GRPO have never been compared side-by-side, leaving practitioners without guidance on the best optimiser for the domain. 
Third, there is no comprehensive benchmark to systematically evaluate financial reasoning across key challenges such as domain-specific language understanding~\cite{xie2024open}, multi-table reasoning~\cite{zhao2024docmath}, long-context document analysis~\cite{reddy2024docfinqa}, and equation-based reasoning~\cite{ma2023xbrl}. Existing benchmarks like PIXIU~\cite{xie2023pixiu}, FinBen~\cite{xie2024finben}, and BizBench~\cite{krumdick2024bizbench} address only parts of this space. Moreover, current studies only evaluated a few LLM's financial reasoning ability.

\begin{table}[h!]
    \centering
    \scriptsize
    \setlength{\tabcolsep}{1.5pt} 
    \renewcommand{\arraystretch}{1.0} 
    \caption{Comparison of Existing Reasoning LLMs and Financial Benchmarks. SFT indicates super-
vised fine-tuning. RL specifies the reinforcement learning method used. CoT data is categorized into Guide, indicate the presence of gold guidelines to direct reasoning generation, IterR means iteratively verification and refinement, DiffF indicates if involving difficulty-aware filtering to emphasize complex cases. Evaluation indicates whether the benchmarks involve single-table or multi-table reasoning, long-context analysis, and equation-based reasoning. ModN indicates the number of models evaluatd in each work. *Indicates works released after Fin-o1.}
    \label{combined_table}
    \begin{tabular}{l|c|cccc|cc|ccccc}
    \toprule
    \multirow{2}{*}{\textbf{Model / Benchmark}} & \multirow{2}{*}{\textbf{Domain}} & \multicolumn{4}{c|}{\textbf{CoT Data}} & \multicolumn{2}{c|}{\textbf{Model}} & \multicolumn{5}{c}{\textbf{Evaluation}} \\
    \cmidrule(lr){3-6} \cmidrule(lr){7-8} \cmidrule(lr){9-13}
    & & \textbf{Public} & \textbf{Guide} & \textbf{IterR} & \textbf{CompF} & \textbf{SFT} & \textbf{RL} & \textbf{Single} & \textbf{Multi} & \textbf{Long Context} & \textbf{Equations} & \textbf{ModN}\\
    \midrule
    PIXIU~\cite{xie2023pixiu} & Financial & $\times$ & $\times$ & $\times$ & $\times$ & $\checkmark$ & $\times$ & $\checkmark$ & $\times$ & $\times$ & $\times$ & 9 \\
    FinBen~\cite{xie2024finben} & Financial & $\times$ & $\times$ & $\times$ & $\times$ & $\times$ & $\times$ & $\checkmark$ & $\times$ & $\times$ & $\times$ & 21 \\ 
    BizBench~\cite{krumdick2024bizbench} & Financial & $\times$ & $\times$ & $\times$ & $\times$ & $\times$ & $\times$ & $\checkmark$ & $\checkmark$ & $\times$ & $\times$ & 16 \\ 
    DocFinQA~\cite{reddy2024docfinqa} & Financial & $\times$ & $\times$ & $\times$ & $\times$ & $\times$ & $\times$ & $\checkmark$ & $\times$ & $\checkmark$ & $\times$ & 12 \\ 
    S1-32B~\cite{muennighoff2025s1} & General & $\checkmark$ & $\times$ & $\times$ & $\checkmark$ & $\checkmark$ & $\times$ & $\times$ & $\times$ & $\times$ & $\checkmark$ & 12 \\ 
    Limo~\cite{ye2025limo} & General & $\checkmark$ & $\times$ & $\times$ & $\checkmark$ & $\checkmark$ & $\times$ & $\times$ & $\times$ & $\times$ & $\checkmark$ & 6 \\ 
    HuatuoGPT-o1~\cite{chen2024huatuogpt} & Other & $\checkmark$ & $\times$ & $\checkmark$ & $\times$ & $\checkmark$ & PPO & $\times$ & $\times$ & $\times$ & $\times$ & 10 \\ 
    DeepSeek-R1~\cite{guo2025deepseek} & General & $\times$ & $\checkmark$ & $\times$ & $\times$ & $\checkmark$ & GRPO & $\times$ & $\times$ & $\times$ & $\times$ & $\times$ \\   
    OpenAI-o1~\cite{jaech2024openai} & General & $\times$ & $\times$ & $\times$ & $\times$ & $\times$ & $\times$ & $\times$ & $\times$ & $\times$ & $\times$ & $\times$ \\ 
    Fin-R1\cite{finr1}\textsuperscript{*} & Financial & $\times$ & $\times$ & $\times$ & $\times$ & $\checkmark$ & GRPO & $\checkmark$ & $\times$ & $\times$ & $\times$ & 10 \\
    DianJin-R1\cite{zhu2025dianjin}\textsuperscript{*} & Financial & $\checkmark$ & $\times$ & $\times$ & $\times$ & $\checkmark$ & GRPO & $\checkmark$ & $\times$ & $\times$ & $\times$ & 12 \\
    \textbf{Fin-o1 (Ours)} & Financial & $\checkmark$ & $\checkmark$ & $\checkmark$ & $\checkmark$ & $\checkmark$ & GRPO, PPO, DPO & $\checkmark$ & $\checkmark$ & $\checkmark$ & $\checkmark$ & 29 \\ 
    \bottomrule
    \end{tabular}
\end{table}

To address these gaps, we present three key contributions. 
First, we curate \textbf{FinCoT}, the first high-quality, publicly available financial CoT dataset built by a novel three-stage framework that combines domain-aware supervision, iterative LLM refinement, and difficulty-aware filtering. 
We construct FinCoT from diverse sources including {FinQA}~\cite{chen2022finqadatasetnumericalreasoning}, 
{ConvFinQA}~\cite{chen2022convfinqa}, {DocFinQA}~\cite{reddy2024docfinqa}, 
{TATQA}~\cite{zhu2021tatqaquestionansweringbenchmark}, 
{Econ\_Logic} \cite{quan2024econlogicqa},
{DocMath-Eval}~\cite{zhao2024docmath}, and {BizBench-QA}~\cite{krumdick2024bizbench}. 
Our new framework yields high-fidelity financial CoT data that improves 8B and 14B models beyond SOTA reasoning models such as GPT-o1, DeepSeek-R1, and GPT-4.5.
Second, we develop \textbf{Fin-o1} (including Fin-o1-8B and Fin-o1-14) based on FinCoT, the first open-source financial reasoning large language models, trained via a two-phase framework combining supervised fine-tuning and RL-based reasoning enhancement. 
Fin-o1 consistently outperform existing financial reasoning models and SOTA general-reasoning LLMs such as GPT-o1, DeepSeek-R1, GPT-4.5.
We systematically investigate the effectiveness of PPO~\cite{schulman2017proximal}, DPO~\cite{rafailov2023direct}, and GRPO~\cite{tang2024generalized} in improving domain-specific reasoning, offering the first empirical study on RL methods for financial reasoning.
Third, we introduce \textbf{FinReason}, the first comprehensive benchmark designed to systematically evaluate the financial reasoning capabilities of LLMs. FinReason covers most LLMs with diverse tasks involving domain-specific language, structured data, and long-context documents, providing a robust framework on datasets such as FinQA~\cite{chen2021finqa}, DM-Simplong~\cite{zhao2024docmath}, XBRL-Math~\cite{wang2025finnlp}, and DM-Complong~\cite{zhao2024docmath}. We conduct a comprehensive evaluation on FinReason, covering 29 large language models from general-purpose, reasoning-enhanced, and financial-specific families, including GPT, LLaMA, DeepSeek, Qwen, and our Fin-o1 models.

Our results reveal that general reasoning models, despite their strong performance on generic benchmarks, struggle with financial reasoning tasks that require precise domain understanding, structured data interpretation, and long-context document reasoning. Frontier financial reasoning LLMs such as Dianjin-R1 and FinR1 exhibit performance degradation compared to their backbones in long-context tasks, highlighting their limited robustness. In contrast, our Fin-o1 models, trained on the FinCoT dataset, consistently outperform their backbone models and surpass larger general-purpose models, demonstrating the importance of domain-specific reasoning supervision. Further analysis shows that reinforcement learning methods have varying effectiveness in this context. While PPO and DPO fail to yield stable improvements, GRPO achieves consistent gains by aligning optimization with task-specific reasoning objectives. These findings illustrate that financial reasoning poses unique challenges that cannot be addressed by scaling model size or applying general reasoning techniques alone, emphasizing the need for specialized data, models, and optimization strategies.

\section{FinCoT Reasoning Training Data}
\begin{figure}
    \centering
    \includegraphics[width=0.9\linewidth]{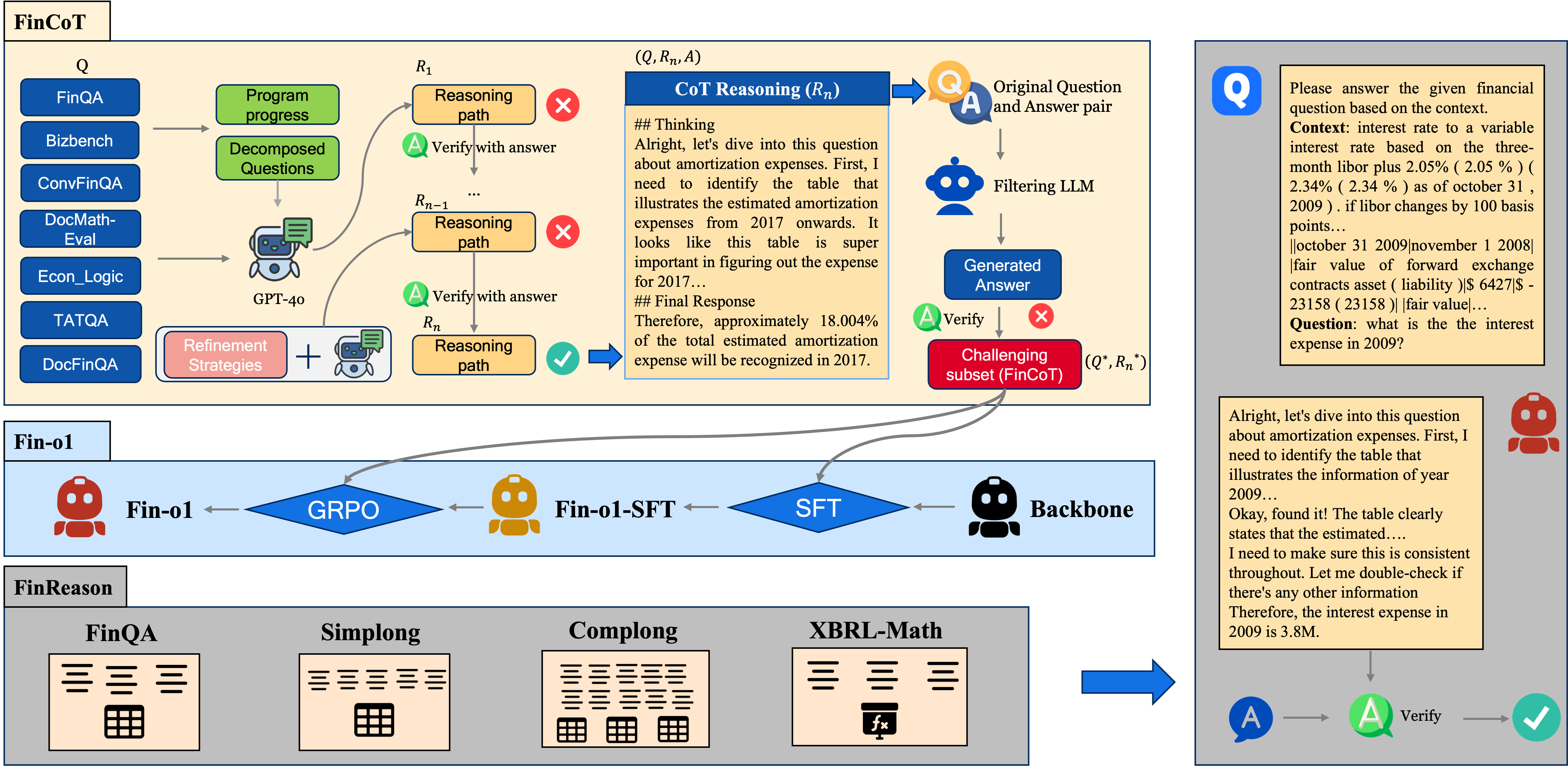}
    \caption{Overall framework of development of FinCoT, Fin-o1 and FinReason.}
    \label{fig:dataset}
\end{figure}

In this section, we describe the construction of FinCoT, the first high-fidelity  dataset of chain-of-thought reasoning paths designed to enhance and evaluate LLMs’ financial reasoning capabilities, as illustrated in Fig. \ref{fig:dataset}.

\subsection{Data Collection}


To develop a robust chain-of-thought training dataset tailored for financial reasoning, we systematically curate data from diverse financial data sources. 
Distinguished from FinR1 \cite{finr1} and Dianjin \cite{zhu2025dianjin}, which only include FinQA for reasoning tasks while the rest of their datasets, i.e, Finance-Instruct \cite{flowers2025finance}, FinCorpus \cite{duxiaoman2023fincorpus}, and others that focus primarily on financial knowledge enhancement\footnote{Samples of these data can be found in Appendix \ref{other_data}}, our dataset explicitly provides structured chain-of-thought supervision to enable complex financial reasoning, each contributing distinct reasoning skills\footnote{See Appendix~\ref{appendix:datasets} for dataset details.}:
\begin{itemize}
    \item FinQA \cite{chen2022finqadatasetnumericalreasoning} and ConvFinQA \cite{chen2022convfinqa}: Focused on numerical reasoning over short financial texts and single well-structured tables, supporting the model in basic arithmetic and aggregation tasks.
    \item TATQA \cite{zhu2021tatqaquestionansweringbenchmark}: Provides semi-structured financial reports with single-table QA tasks, enhancing the model’s ability to reason over less curated, real-world financial tables.
    \item DocMath-Eval \cite{zhao2024docmath}: Adds mathematical reasoning over structured tables, introducing complex quantitative reasoning patterns.
    \item Econ-Logic \cite{quan2024econlogicqa}: Includes financial decision-making and logical inference tasks, requiring models to apply domain-specific logic beyond numerical calculation.
    \item BizBench-QA \cite{krumdick2024bizbench}: Introduces multi-table reasoning tasks, training the model to integrate and synthesize information across multiple data sources.
    \item DocFinQA \cite{reddy2024docfinqa}: Comprises long-form financial documents with QA tasks, enabling the model to handle extended context and multi-step reasoning over lengthy narratives.
\end{itemize}

\subsection{Iterative Chain-of-Thought Generation and Verification with Domain-Guided Supervision}
\label{reasoning}
Building our raw data collection, 
we further curate the first dataset containing 9,186 chain-of-thought annotated QA pairs, specifically designed to support SFT and RL for financial reasoning models, via a three-stage CoT generation framework by integrating domain-aware supervision, iterative refinement, and difficulty-aware filtering. This framework addresses the limitations of existing chain-of-thought methods, which prompt LLMs such as DeepSeek-R1 or Gemini \cite{team2024gemini, ye2025limo, muennighoff2025s1} to generate reasoning paths in a single round \cite{finr1, zhu2025dianjin}, sufficient for general tasks but inadequate for financial reasoning which requires precise, multi-step, domain-informed calculations. For example, financial tasks like calculating profits require carefully following each calculation step with the right data and rules, where even small mistakes can completely invalidate the final answer.

\begin{figure}
    \centering
    \includegraphics[width=0.5\linewidth]{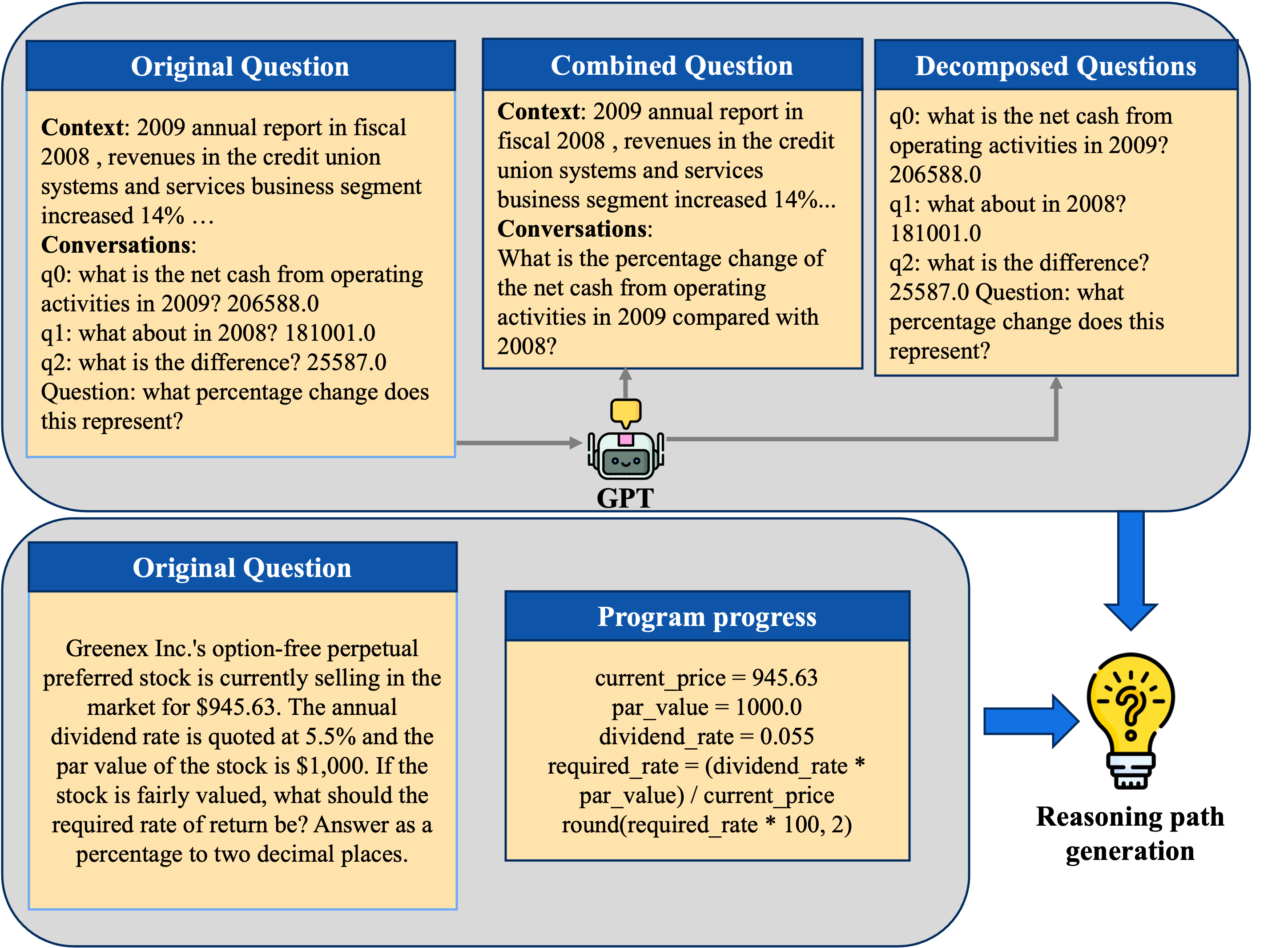}
    \caption{Workflow of curating the combined question and examples of program progress.}
    \label{fig:decompose}
\end{figure}

In the first stage, we tackle the lack of structured guidance in existing methods by incorporating programmatic steps and expert-curated sub-questions from datasets such as BizBench and ConvFinQA, collected during our data preparation phase. While ConvFinQA offers high-quality sub-questions, it does not provide complete question-answer pairs that capture the full financial decision-making process. To close this gap, we use GPT-4o to synthesize these sub-questions into combined, holistic questions (see Fig. \ref{fig:decompose}. This allows the model to see both the detailed steps and how they fit into a broader financial reasoning workflow, helping it generate more coherent and logically consistent reasoning paths that mirror domain expert expectations).

In the second stage, we address the brittleness of single-pass reasoning by introducing an iterative verification and refinement process, enabling the model to correct its reasoning progressively. Inspired by HuatuoGPT-o1 \cite{chen2024huatuogpt} but adapted to the stricter demands of financial reasoning, we prompt GPT-4o to generate step-by-step reasoning paths, which are independently evaluated by a verifier (also GPT-4o). Instead of discarding incorrect outputs entirely, our framework forces the model to revisit earlier steps, refine its assumptions, and adjust its reasoning flow based on verifier feedback (see Fig. \ref{fig:refine} in appendix \ref{app:example}.). This allows the model to detect and fix errors that might otherwise propagate unchecked, ensuring greater accuracy and robustness, especially in complex financial tasks where early missteps can have outsized impacts.

In the third stage, we introduce a difficulty-aware filtering step to ensure that the final dataset targets genuinely challenging reasoning scenarios (as shown in Fig. \ref{fig:dataset}). Unlike prior methods \cite{chen2024huatuogpt, finr1} that focus only on verifying correctness, we also consider question difficulty, motivated by recent studies showing that easy QA pairs do little to advance reasoning ability \cite{ye2025limo, muennighoff2025s1}. We use \texttt{Llama3.1-8B-Instruct} to attempt each question, then apply a verifier (see Section \ref{evaluation_setting}) to check the answers. QA pairs that are correctly solved by the filtering LLM are excluded, which is deemed less challenging. 



\section{Fin-o1}
In this section, we propose Fin-o1 (including Fin-o1-8B and Fin-o1-14B) based on FinCoT, the first open source LLMs targeted on financial reasoning tasks. \textbf{Qwen3} are selected as backbone models, given their strong performance among small and mid-sized LLMs (shown in Table \ref{tab:llm_performance_full}).
Following previous studies~\cite{chen2024huatuogpt,dubey2024llama}, a two-stage training approach, supervised fine-tuning and RL, is adopted to develop the model.  
FinCoT is divided into two parts FinCoT\_SFT (7686) and FinCoT\_RL (1,500) to support the training. Based on FinCoT, we further conduct the first evaluation of different RL methods to assess their impact on financial reasoning tasks. 

\subsection{Supervised fine-tuning}
In the first stage, we perform \textit{supervised fine-tuning}, with the FinCoT\_SFT, aiming to guide the model to systematically structure its reasoning process before providing the final answer. As shown in the right part of Fig. \ref{fig:dataset}, each training instance is composed of a prompt, contextual information extracted from textual and tabular data, and a question that requires multi-step reasoning. The output target consists of a detailed reasoning path, followed by a summarized answer, emphasizing the importance of step-by-step logical analysis before deriving the final response. By structuring the output in this manner, the objective of SFT is to cultivate a disciplined reasoning framework, encouraging the model to first articulate a coherent analytical process, thereby enhancing interpretability and accuracy in complex financial reasoning scenarios. After the supervised fine-tuning, we get Fin-o1-SFT.

\subsection{Reinforcement learning}
In the second stage, we conduct a comprehensive evaluation of various RL methods to assess their impact on financial reasoning tasks. 
To effectively evaluate these methods, we specially design targeted reward functions and curate high-quality positive and negative samples tailored to each method. This comparative analysis aims to provide insights into the strengths and limitations of each RL method in enhancing reasoning capabilities for complex financial tasks.


\paragraph{Proximal policy optimization.}

We start from PPO \cite{schulman2017proximal}. It optimizes the policy by maximizing the expected reward while constraining the extent of policy updates, thus preventing excessive divergence from the reference policy.
Each input \( x \) is processed by the Fin-o1-SFT to generate a response with a reasoning path \( y^* \). The reward \( r \) is defined based on the verifier feedback when comparing if the \( y^* \) is consistent with the gold answer \( y \). To effectively evaluate the generated answer with the gold, GPT-4o is used to verify the correctness of the results. 
\begin{equation}
r(x, y, y^*) = 
\begin{cases} 
1 & \text{if } \text{verifier}(y, y^*) = \text{True} \\
0 & \text{otherwise}
\end{cases}
\end{equation}

Additionally, the reward is adjusted by incorporating the Kullback-Leibler (KL) divergence penalty between the updated policy \( \pi_{\theta} \) and the reference policy \( \pi_{ref} \) \cite{luong2024reft}, aiming to prevent excessive deviation from the reference policy while ensuring stable learning.:

\begin{equation}
r'(x, y, y^*) = r(x, y, y^*) - \beta \cdot KL(\pi_{\theta} \parallel \pi_{ref})
\end{equation}





\paragraph{Direct preference optimization. }

DPO \cite{rafailov2023direct} refines the model’s decision-making by systematically contrasting positive and negative samples through pairwise comparisons. In our framework, we extend the conventional DPO setup by implementing a more strategically designed sampling mechanism tailored to the nuances of complex financial reasoning paths.

The iterative CoT reasoning path generation process is leveraged in this step to effectively curate high-quality positive-negative samples. As detailed in \ref{reasoning}, this process involves employing GPT-4o to generate reasoning paths that undergo iterative verification and correction. Rather than employing low-quality negative samples that diverge significantly from the positive samples, we implement a targeted selection strategy that designates the last failed reasoning path \( R_{n-1} \) as the negative sample. This path is specifically chosen as it closely mirrors the input query in logical structure while containing identifiable reasoning flaws, making it a more informative negative sample. 
By focusing on plausibility and logical coherence, our sampling strategy mitigates the risk of selecting trivial or overly simplistic negative samples, thereby reinforcing the model’s capacity to differentiate between nuanced reasoning paths and effectively learn from more challenging and contextually relevant negative samples.




\paragraph{Generalized proximal policy optimization. }

Compared with PPO that optimizes the model by maximizing expected rewards through constrained policy updates, GRPO~\cite{tang2024generalized} extends PPO by incorporating additional objectives or multi-task learning capabilities, making it suited for financial reasoning tasks where accuracy and reasoning completeness, long-context reasoning abilities are crucial for effectively handling complex financial scenarios.

Therefore, to comprehensively improve LLMs' reasoning abilities in financial reasoning tasks, we propose a novel multi-faceted reward structure in GRPO that systematically evaluates four critical dimensions. Compared with prior work that predominantly targets accuracy and format \cite{chen2024huatuogpt, finr1, zhu2025dianjin}, our framework introduces two additional components: reasoning logic and length rewards. By integrating these dimensions, we aim to mitigate shortcut learning, reinforce logical consistency, and incentivize effective reasoning over long contexts—a crucial capability given the inherently lengthy and complex nature of financial documents. Formally, the reward function for GRPO is designed as follows:

\begin{equation}
r(x, y, y^*) = \alpha_1 \cdot r_{acc} + \alpha_2 \cdot r_{logic} + \alpha_3 \cdot r_{format} + \alpha_4 \cdot r_{length} \cdot r_{acc}
\end{equation}

where:

\begin{itemize}
    \item \( r_{acc} \): Accuracy reward, based on the verifier’s assessment of the final response correctness.
    \item \( r_{logic} \): Logical consistency reward, evaluating the correctness of the reasoning path.
    \item \( r_{format} \): Assessing if the generated response follows the designed format.
    \item \( r_{length} \): To encourage the model’s ability to reason over lengthy contexts—a critical requirement in financial analysis given the extensive nature of many financial documents—we provide an additional reward when the answer is correct and the context length exceeds 8,192 tokens. If the accuracy reward is zero, the length reward is also set to zero.
\end{itemize}

To effectively evaluate the generated response, we also leverage LLMs (GPT-4o) as judge. Following the reward score design in \cite{finr1,chen2024huatuogpt}, we first assesses the accuracy of the generated response, signing a reward of 1 if the answer is deemed correct. Similarly, if the logical reasoning is validated as correct, a reward of 1 is also given. For format and length evaluation, we employ rule-based methods to assign corresponding rewards, the format reward is set as 0.1 and the length reward is set as 1. 







\section{FinReason Evaluation Benchmark}
In this section, we present \textbf{FinReason}, the first comprehensive benchmark listed in Table \ref{tab:datasets} designed to systematically evaluate the financial reasoning capabilities of LLMs.
\subsection{Tasks}
Financial reasoning tasks present distinct challenges beyond general reasoning, as they require not only understanding the structure and semantic content of financial documents and tables but also accurately interpreting and executing complex equations.
To this end, we introduce a unified benchmark that systematically integrates reasoning over \textbf{(1) financial texts with tables, (2) long-context reasoning, (3) multi-table analysis, and (4) interpretation and execution of financial equations}, establishing a comprehensive evaluation framework to assess and elevate LLM capabilities in financial reasoning tasks.
Four datasets are selected to evaluate LLMs across these key aspects of financial reasoning:

\textbf{(1) FinQA}~\cite{chen2021finqa} serves as the foundational benchmark for quantitative reasoning over structured tables and basic financial texts. It assesses the model’s ability to extract and synthesize financial information related to core financial concepts such as revenue, profit margins, and financial ratios. 

\textbf{(2) Simplong}~\cite{zhao2024docmath} targets long-context reasoning within single-table scenarios. It assesses the model’s capacity to perform numerical reasoning based on structured data while maintaining context throughout extended financial texts. 

\textbf{(3) Complong}~\cite{zhao2024docmath} extends Simplong by introducing multi-table scenarios, challenging the model to integrate information across multiple tables. This dataset tests the model’s ability to handle more complex financial reporting contexts, where financial data is fragmented and requires cross-table reasoning to draw accurate conclusions.

\textbf{(4) XBRL-Math}~\cite{wang2025finnlp} emphasizes equation-based reasoning, requiring the model to accurately interpret and execute financial formulas derived from structured data in XBRL filings. 
This dataset assesses the model’s ability to associate financial terms with corresponding numerical data and perform structured calculations.

\begin{table}[h]
    \centering
    \footnotesize
    \setlength{\tabcolsep}{2pt} 
    \caption{Overview of the datasets evaluated in the study.}
    \label{tab:datasets}
    \begin{tabular}{lr c cr c c}  
        \toprule
        \textbf{Dataset} & \textbf{Size} & \textbf{Data Types} & \textbf{Data Source} & \textbf{Avg. Token} & \textbf{Metric} &\textbf{License} \\
        \midrule
        FinQA~\cite{chen2021finqa} & 1,100 & Tables and Texts & Financial reports &1,128 & Acc &  MIT License\\
        DM-Simplong~\cite{zhao2024docmath} & 100 & Tables and Texts & Financial reports &4,330 & Acc &  MIT License\\
        DM-Complong~\cite{zhao2024docmath} & 300 & Tables and Texts & Financial reports &39,983& Acc &  MIT License\\
        XBRL-Math~\cite{wang2025finnlp} & 90 & Texts and Equations & \makecell{XBRL}  &397 & Acc & CDLA-Permissive 2.0 \\
        \bottomrule
    \end{tabular}
\end{table}


\subsection{Evaluated Models}
To investigate how existing LLMs, especially reasoning-enhanced LLMs, perform in financial tasks,  we select a diverse set of 29 LLMs from the \textbf{GPT}, \textbf{LLaMA}, \textbf{DeepSeek}, and \textbf{Qwen} families, along with other reasoning models including both general domain and financial specific models. Our selection covers models of various scales, from smaller models (8B) to large-scale models (70B), and even ultra-large models like \textbf{DeepSeek-R1} (671B). We also include advanced closed-source models such as \textbf{GPT-4o} and \textbf{GPT-o1} as strong baselines.
More details of the models can be found in Appendix \ref{model_description}.

\subsection{Evaluation Settings}
\label{evaluation_setting}
In each dataset, LLMs are presented with a context containing financial reports, tables, equations, and symbols, along with a question. Following prior work \cite{wei2022chain}, we use a generic prompt (see appendix \ref{app:prompt} for more details) to instruct models to answer the question. 

While answers to financial tasks are often expressed in mathematical forms, variations in representation, such as differences in percentages, rounding conventions, or numeric formatting, can lead to discrepancies in surface-level comparisons.
To address this, we opt out exact match metrics and instead adopt an \textbf{LLM-as-judge} evaluation paradigm. Specifically, we follow the answer extraction methodology proposed by \cite{chen2023theoremqa} in TheoremQA, extracting the numerical results from model outputs. The extracted answers are then compared against the ground truth to assess correctness, allowing for flexible yet rigorous evaluation of generated responses. Please see appendix \ref{model_evaluation} for more details of the evaluation setting.

\section{Results}
Next, we present the results of these LLMs on FinReason and discuss our main findings through the results for the research questions listed in section ~\ref{intro}.

\begin{table}[ht]
    \centering
    \tiny
    \caption{Overall performance of different LLMs on four financial reasoning datasets.}
    \renewcommand{\arraystretch}{1}
    \resizebox{0.8\columnwidth}{!}{%
    \begin{tabular}{lccccc}
        \toprule
        \textbf{Models} & \textbf{FinQA} & \textbf{DM-Simplong} & \textbf{XBRL-Math} & \textbf{DM-Complong} & \textbf{Average} \\
        \midrule
        GPT-4o & 72.49 & \textbf{60.00} & 72.22 & 39.33 & 61.01 \\
        GPT-o1-preview & 49.07 & 56.00 & 74.44 & 36.67 & 54.05 \\
        GPT-o3-mini & 60.87 & 59.00 & 76.67 & 35.00 & 57.89 \\
        DeepSeek-V3 & 73.20 & 53.00 & 76.67 & \textbf{42.33} & \textbf{61.30} \\
        DeepSeek-R1 & 65.13 & 53.00 & 86.67 & 38.67 & 60.87 \\
        GPT-4.5 & 68.94 & 59.00 & 74.44 & 39.33 & 60.43 \\
        \midrule
        Llama-4-Scout & 70.45 & 52.00 & \textbf{88.89} & 0.67 & 53.00 \\
        Llama-3-70B-Instruct & 58.92 & 41.00 & 56.67 & 13.67 & 42.57 \\
        Llama-3.1-70B-Instruct & 63.18 & 48.00 & 63.33 & 34.33 & 52.21 \\
        Llama-3.3-70B-Instruct & 68.15 & 54.00 & 70.00 & 32.00 & 56.04 \\
        Qwen2.5-72B-Instruct & {73.38} & 59.00 & 67.78 & 14.67 & 53.71 \\
        Qwen2.5-Math-72B-Instruct & 69.74 & 42.00 & 83.33 & 5.00 & 50.02 \\
        DeepSeek-R1-Distill-Llama-70B & 66.73 & 53.00 & 86.67 & 30.67 & 59.27 \\
        \midrule
        Qwen2.5-32B-Instruct & 73.11 & 56.00 & 65.56 & 30.00 & 56.17 \\
        Qwen3-32B & 64.15 & 51.00 & 85.56 & 26.00 & 56.68 \\
        QwQ-32B & 61.22 & 46.00 & 84.44 & 20.00 & 52.92 \\
        DeepSeek-R1-Distill-Qwen-32B & 65.48 & 55.00 & 84.44 & 24.67 & 57.40 \\
        Limo & 63.44 & 45.00 & 61.11 & 15.33 & 46.22 \\
        S1-32B & 66.81 & 53.00 & 84.44 & 24.00 & 57.06 \\

        \midrule
        Qwen2.5-14B-Instruct & 67.44 & 59.00 & 57.78 & 26.67 & 52.72 \\
        Qwen3-14B & 64.33 & 49.00 & 86.67 & 24.00 & 56.00 \\
        DeepSeek-R1-Distill-Qwen-14B & 63.27 & 44.00 & 84.44 & 21.00 & 53.18 \\
        DeepSeek-R1-Distill-Llama-8B & 45.96 & 33.00 & 81.11 & 15.67 & 43.94 \\
        Llama-3-8B-Instruct & 41.97 & 29.00 & 48.89 & 6.00 & 31.47 \\
        Llama-3.1-8B-Instruct & 54.13 & 34.00 & 62.22 & 14.30 & 41.16 \\
        
        Qwen2.5-7B-Instruct & 55.37 & 41.00 & 42.22 & 17.67 & 39.07 \\
        Qwen3-8B & 62.11 & 46.00 & 83.33 & 17.67 & 52.28 \\
        FinR1-7B & 58.74 & 37.00 & 30.00 & 13.67 & 34.85 \\
        Dianjin-R1-7B & 60.20 & 35.00 & 83.33 & 14.67 & 48.30 \\
        \midrule
    
        Fino1-8B & 73.03 & 56.00 & 84.44 & 26.33 & 59.95 \\
        Fino1-14B & \textbf{74.18} & 55.00 & {87.78} & 27.33 & 61.07\\
        \bottomrule
    \end{tabular}
    }
    \label{tab:llm_performance_full}
\end{table}

Table~\ref{tab:llm_performance_full} presents the performance of evaluated LLMs across four financial reasoning benchmarks: FinQA, DM-Simplong, XBRL-Math, and DM-Complong.
Our \textbf{Fin-o1-14B} achieves a score of 61.07, ranking second only to Deepseek-V3 despite having substantially fewer parameters. Notably, Fin-o1-14B surpasses larger-scale models like LLaMA-3.3-70B-Instruct, Qwen2.5-72B-Instruct, even GPT-o1-preview, GPT-o3-mini, and GPT-4.5, underscoring its superior financial reasoning capabilities.
\textbf{Fin-o1-8B} also performs competitively, attaining an average score of 58.58, exceeding most models with 70B parameters or fewer. Both Fin-o1-8B and Fin-o1-14B excel in FinQA, DM-Simplong, and XBRL-Math, demonstrating the impact of targeted domain adaptation in financial reasoning tasks.
These gains also confirm the effectiveness of our three-stage FinCoT construction and the two-phase SFT + RL training strategy.


\paragraph{General reasoning LLMs fail in financial reasoning.} 
While general reasoning models exhibit significant improvements in general mathematical reasoning tasks, they struggle to effectively transfer these gains to financial reasoning tasks, which demand specialized knowledge and reasoning over financial documents and tables.  
For instance, the performance of QwQ-32B and Limo declined notably compared to Qwen2.5-32B-Instruct, dropping from 56.17\% to 52.92\% and 46.22\%, respectively. Similarly, although S1-32B demonstrated overall performance gains, its effectiveness in \textit{FinQA}, \textit{Simplong} and \textit{Complong} decreased. These results highlight the persistent gap between general and financial reasoning due to the complexities of interpreting financial tables and structures.

\textbf{Incomprehensive reasoning data limits consistent improvements.} While our model demonstrates consistent improvements over the backbone model Qwen3 across all datasets, other financial models do not exhibit such stable improvements. 
For instance, while Fin-R1 and Dianjin-R1-7B achieved higher scores in \textit{FinQA}, their performance in long-context reasoning tasks dropped significantly compared with their backbone model Qwen-2.5-7B-Instruct. On \textit{Simplong}, the backbone model scored 41\%, but Fin-R1 and Dianjin-R1-7B fell to 37\% and 35\%, respectively. A similar decline is observed on \textit{Complong}, indicating the challenges in maintaining long context reasoning performance.
These declines are attributable to CoT datasets used in these works, which focused on shorter contexts and knowledge-based QA rather than reasoning over contexts, limiting the models' exposure to reasoning over extended financial contexts.


\begin{table}[ht]
    \centering
    \caption{Ablation study of Fin-o1-8B with different SFT training data and RL methods.}
    \renewcommand{\arraystretch}{1.1}
    \setlength{\tabcolsep}{4pt}
    \resizebox{0.7\columnwidth}{!}{%
    \begin{tabular}{lccccc}
        \toprule
        \textbf{Models} & \textbf{FinQA} & \textbf{DM-Simplong} & \textbf{XBRL-Math} & \textbf{DM-Complong} & \textbf{Average} \\
        \midrule
        \multicolumn{6}{l}{\textit{Backbone model}} \\
        Qwen3-8B & 62.11 & 46.00 & 83.33 & 17.67 & 52.28 \\
        \midrule
        \multicolumn{6}{l}{\textit{SFT model}} \\
        Fin-o1-8B (SFT with original QA pair) & 66.55 & 50.00 & 74.44 & 16.67 & 51.92 \\
        Fin-o1-8B (SFT with all CoT data) & 71.32 & 51.00 & 81.11 & 21.67 & 56.28 \\
        Fin-o1-8B (SFT with filtered CoT) & 71.78 & \textbf{57.00} & 82.22 & 23.33 & 58.58 \\
        \midrule
        \multicolumn{6}{l}{\textit{SFT + RL model}} \\
        Fin-o1-8B (SFT + PPO) & 72.05 & 54.00 & 83.33 & 23.00 & 58.10 \\
        Fin-o1-8B (SFT + DPO) & 69.74 & 51.00 & 72.22 & 25.00 & 54.49 \\
        Fin-o1-8B (SFT + GRPO) & \textbf{73.03} & 56.00 & \textbf{84.44} & \textbf{26.33} & \textbf{59.95} \\

        \bottomrule
    \end{tabular}
    }
    \label{tab:rl_performance}
\end{table}

Table~\ref{tab:rl_performance} presents additional results evaluating the effectiveness of various RL methods in financial reasoning tasks and further verifying the effectiveness of FinCoT. Based on the results, we find that: 

\paragraph{GRPO Outperforms Other RL Methods in Financial Reasoning.}
GRPO achieves the most substantial gains, driven by its multi-faceted reward function that emphasizes correct reasoning and long-context comprehension. Notably, in DM-Complong, where other models like Fin-R1 and S1-32B falter, the SFT+GRPO model reaches 26.33\% with just 8B parameters—a 9\% improvement over its backbone model.
In contrast, PPO shows gains in \textit{FinQA} and \textit{XBRL-Math} but declines in \textit{Simplong} and \textit{Complong}, indicating weaker long-context reasoning. For DPO, while the performance in DM-Complong improves, performance in simpler tasks declines. This may stem from the design of positive-negative sample pairs, where the negative path is curated from the last-mistake path and the positive path is generated by allowing the model to take one additional reasoning step. This approach may inadvertently encourage excessive reasoning, benefiting complex tasks but potentially causing overthinking in simpler tasks.

We further compare the impact of fine-tuning the backbone model using original QA pairs without CoT reasoning and with unfiltered CoT data.
Training without CoT data slightly improves basic financial reasoning tasks like \textit{FinQA}, but leads to performance drops in \textit{XBRL-Math} and \textit{Complong}. This decline is particularly pronounced in \textit{XBRL-Math}, where the lack of step-by-step reasoning impedes the model’s ability to handle financial equations, hindering its generalization to equation-heavy tasks.
In contrast, using the unfiltered FinCoT dataset demonstrates the effectiveness of exposing the model to more complex reasoning paths, underscoring the importance of reasoning-focused training for financial tasks.



\section{Conclusion}  

In this work, we address the critical yet underexplored challenge of financial reasoning for LLMs. We present three key contributions to bridge existing gaps: (1) the development of FinCoT, the first publicly available CoT dataset specifically curated for financial reasoning tasks; (2) the introduction of Fin-o1, the first open-source financial reasoning model trained through a two-phase framework combining supervised fine-tuning and RL methods; and (3) the establishment of FinReason, a comprehensive benchmark designed to systematically evaluate financial reasoning across essential components such as multi-table analysis, long-context reasoning, and equation-based tasks.

Our extensive evaluations demonstrate that general reasoning models struggle to adapt to financial contexts, revealing significant overall performance drops. In contrast, our Fin-o1 models, trained on FinCoT with GRPO, achieve consistent gains across diverse financial reasoning tasks, highlighting the critical role of domain-specific CoT supervision and structured RL optimization. These findings underscore the need for targeted datasets, specialized models, and advanced training strategies to effectively address the complexities of financial reasoning, setting a new standard for future research in this domain.



    




\newpage
\bibliographystyle{plain}
\bibliography{paper}

\begin{thebibliography}{10}

\bibitem{chen2024huatuogpt}
Junying Chen, Zhenyang Cai, Ke~Ji, Xidong Wang, Wanlong Liu, Rongsheng Wang, Jianye Hou, and Benyou Wang.
\newblock Huatuogpt-o1, towards medical complex reasoning with llms.
\newblock {\em arXiv preprint arXiv:2412.18925}, 2024.

\bibitem{chen2023automatic}
Mark Chen et~al.
\newblock Automatic formal reasoning about large language models.
\newblock {\em arXiv preprint arXiv:2307.09829}, 2023.

\bibitem{chen2023theoremqa}
Wenhu Chen, Ming Yin, Max Ku, Pan Lu, Yixin Wan, Xueguang Ma, Jianyu Xu, Xinyi Wang, and Tony Xia.
\newblock Theoremqa: A theorem-driven question answering dataset.
\newblock {\em arXiv preprint arXiv:2305.12524}, 2023.

\bibitem{chen2021finqa}
Zhiyu Chen, Wenhu Chen, Charese Smiley, Sameena Shah, Iana Borova, Dylan Langdon, Reema Moussa, Matt Beane, Ting-Hao Huang, Bryan Routledge, et~al.
\newblock Finqa: A dataset of numerical reasoning over financial data.
\newblock {\em arXiv preprint arXiv:2109.00122}, 2021.

\bibitem{chen2022finqadatasetnumericalreasoning}
Zhiyu Chen, Wenhu Chen, Charese Smiley, Sameena Shah, Iana Borova, Dylan Langdon, Reema Moussa, Matt Beane, Ting-Hao Huang, Bryan Routledge, and William~Yang Wang.
\newblock Finqa: A dataset of numerical reasoning over financial data, 2022.

\bibitem{chen2022convfinqa}
Zhiyu Chen, Shiyang Li, Charese Smiley, Zhiqiang Ma, Sameena Shah, and William~Yang Wang.
\newblock Convfinqa: Exploring the chain of numerical reasoning in conversational finance question answering.
\newblock {\em arXiv preprint arXiv:2210.03849}, 2022.

\bibitem{drori2023neural}
Iddo Drori et~al.
\newblock Neural-symbolic mathematical reasoning.
\newblock {\em arXiv preprint arXiv:2301.07076}, 2023.

\bibitem{dubey2024llama}
Abhimanyu Dubey, Abhinav Jauhri, Abhinav Pandey, Abhishek Kadian, Ahmad Al-Dahle, Aiesha Letman, Akhil Mathur, Alan Schelten, Amy Yang, Angela Fan, et~al.
\newblock The llama 3 herd of models.
\newblock {\em arXiv preprint arXiv:2407.21783}, 2024.

\bibitem{flowers2025finance}
Joseph~G. Flowers.
\newblock Finance instruct 500k.
\newblock \url{https://huggingface.co/datasets/Josephgflowers/Finance-Instruct-500k}, 2025.
\newblock Accessed: 2025-03-18.

\bibitem{guo2025deepseek}
Daya Guo, Dejian Yang, Haowei Zhang, Junxiao Song, Ruoyu Zhang, Runxin Xu, Qihao Zhu, Shirong Ma, Peiyi Wang, Xiao Bi, et~al.
\newblock Deepseek-r1: Incentivizing reasoning capability in llms via reinforcement learning.
\newblock {\em arXiv preprint arXiv:2501.12948}, 2025.

\bibitem{hurst2024gpt}
Aaron Hurst, Adam Lerer, Adam~P Goucher, Adam Perelman, Aditya Ramesh, Aidan Clark, AJ~Ostrow, Akila Welihinda, Alan Hayes, Alec Radford, et~al.
\newblock Gpt-4o system card.
\newblock {\em arXiv preprint arXiv:2410.21276}, 2024.

\bibitem{jaech2024openai}
Aaron Jaech, Adam Kalai, Adam Lerer, Adam Richardson, Ahmed El-Kishky, Aiden Low, Alec Helyar, Aleksander Madry, Alex Beutel, Alex Carney, et~al.
\newblock Openai o1 system card.
\newblock {\em arXiv preprint arXiv:2412.16720}, 2024.

\bibitem{kou2024automate}
Zhizhuo Kou, Holam Yu, Jingshu Peng, and Lei Chen.
\newblock Automate strategy finding with llm in quant investment.
\newblock {\em arXiv preprint arXiv:2409.06289}, 2024.

\bibitem{krumdick2024bizbench}
Michael Krumdick, Rik Koncel-Kedziorski, Viet~Dac Lai, Varshini Reddy, Charles Lovering, and Chris Tanner.
\newblock Bizbench: A quantitative reasoning benchmark for business and finance.
\newblock In {\em Proceedings of the 62nd Annual Meeting of the Association for Computational Linguistics (Volume 1: Long Papers)}, pages 8309--8332, 2024.

\bibitem{kwon2023efficient}
Woosuk Kwon, Zhuohan Li, Siyuan Zhuang, Ying Sheng, Lianmin Zheng, Cody~Hao Yu, Joseph Gonzalez, Hao Zhang, and Ion Stoica.
\newblock Efficient memory management for large language model serving with pagedattention.
\newblock In {\em Proceedings of the 29th Symposium on Operating Systems Principles}, pages 611--626, 2023.

\bibitem{lewkowycz2022solving}
Aitor Lewkowycz et~al.
\newblock Solving quantitative reasoning problems with language models.
\newblock {\em arXiv preprint arXiv:2206.14858}, 2022.

\bibitem{liu2024deepseek}
Aixin Liu, Bei Feng, Bing Xue, Bingxuan Wang, Bochao Wu, Chengda Lu, Chenggang Zhao, Chengqi Deng, Chenyu Zhang, Chong Ruan, et~al.
\newblock Deepseek-v3 technical report.
\newblock {\em arXiv preprint arXiv:2412.19437}, 2024.

\bibitem{liu2025findabench}
Shu Liu, Shangqing Zhao, Chenghao Jia, Xinlin Zhuang, Zhaoguang Long, Jie Zhou, Aimin Zhou, Man Lan, and Yang Chong.
\newblock Findabench: Benchmarking financial data analysis ability of large language models.
\newblock In {\em Proceedings of the 31st International Conference on Computational Linguistics}, pages 710--725, 2025.

\bibitem{finr1}
Zhaowei Liu, Xin Guo, Fangqi Lou, Lingfeng Zeng, Jinyi Niu, Zixuan Wang, Jiajie Xu, Weige Cai, Ziwei Yang, Xueqian Zhao, et~al.
\newblock Fin-r1: A large language model for financial reasoning through reinforcement learning.
\newblock {\em arXiv preprint arXiv:2503.16252}, 2025.

\bibitem{luong2024reft}
Trung~Quoc Luong, Xinbo Zhang, Zhanming Jie, Peng Sun, Xiaoran Jin, and Hang Li.
\newblock Reft: Reasoning with reinforced fine-tuning.
\newblock {\em arXiv preprint arXiv:2401.08967}, 3, 2024.

\bibitem{ma2023xbrl}
Wei Ma et~al.
\newblock Xbrl-math: A dataset for mathematical reasoning in financial reports.
\newblock {\em Conference on Financial Technology and Natural Language Processing}, 2023.

\bibitem{muennighoff2025s1}
Niklas Muennighoff, Zitong Yang, Weijia Shi, Xiang~Lisa Li, Li~Fei-Fei, Hannaneh Hajishirzi, Luke Zettlemoyer, Percy Liang, Emmanuel Cand{\`e}s, and Tatsunori Hashimoto.
\newblock s1: Simple test-time scaling.
\newblock {\em arXiv preprint arXiv:2501.19393}, 2025.

\bibitem{ouyang2022training}
Long Ouyang et~al.
\newblock Training language models to follow instructions with human feedback.
\newblock {\em arXiv preprint arXiv:2203.02155}, 2022.

\bibitem{peng2025plutus}
Xueqing Peng, Triantafillos Papadopoulos, Efstathia Soufleri, Polydoros Giannouris, Ruoyu Xiang, Yan Wang, Lingfei Qian, Jimin Huang, Qianqian Xie, and Sophia Ananiadou.
\newblock Plutus: Benchmarking large language models in low-resource greek finance.
\newblock {\em arXiv preprint arXiv:2502.18772}, 2025.

\bibitem{quan2024econlogicqa}
Yinzhu Quan and Zefang Liu.
\newblock Econlogicqa: A question-answering benchmark for evaluating large language models in economic sequential reasoning.
\newblock {\em arXiv preprint arXiv:2405.07938}, 2024.

\bibitem{rafailov2023direct}
Rafael Rafailov, Archit Sharma, Eric Mitchell, Christopher~D Manning, Stefano Ermon, and Chelsea Finn.
\newblock Direct preference optimization: Your language model is secretly a reward model.
\newblock {\em Advances in Neural Information Processing Systems}, 36:53728--53741, 2023.

\bibitem{reddy2024docfinqa}
Varshini Reddy, Rik Koncel-Kedziorski, Viet~Dac Lai, Michael Krumdick, Charles Lovering, and Chris Tanner.
\newblock Docfinqa: A long-context financial reasoning dataset.
\newblock {\em arXiv preprint arXiv:2401.06915}, 2024.

\bibitem{schulman2017proximal}
John Schulman, Filip Wolski, Prafulla Dhariwal, Alec Radford, and Oleg Klimov.
\newblock Proximal policy optimization algorithms.
\newblock {\em arXiv preprint arXiv:1707.06347}, 2017.

\bibitem{srivastava2024evaluating}
Pragya Srivastava, Manuj Malik, Vivek Gupta, Tanuja Ganu, and Dan Roth.
\newblock Evaluating llms’ mathematical reasoning in financial document question answering.
\newblock In {\em Findings of the Association for Computational Linguistics ACL 2024}, pages 3853--3878, 2024.

\bibitem{tang2024generalized}
Yunhao Tang, Zhaohan~Daniel Guo, Zeyu Zheng, Daniele Calandriello, R{\'e}mi Munos, Mark Rowland, Pierre~Harvey Richemond, Michal Valko, Bernardo~{\'A}vila Pires, and Bilal Piot.
\newblock Generalized preference optimization: A unified approach to offline alignment.
\newblock {\em arXiv preprint arXiv:2402.05749}, 2024.

\bibitem{duxiaoman2023fincorpus}
Duxiaoman~DI Team.
\newblock Fincorpus.
\newblock \url{https://huggingface.co/datasets/Duxiaoman-DI/FinCorpus}, 2023.
\newblock Accessed: 2024-03-18.

\bibitem{team2024gemini}
Gemini Team, Petko Georgiev, Ving~Ian Lei, Ryan Burnell, Libin Bai, Anmol Gulati, Garrett Tanzer, Damien Vincent, Zhufeng Pan, Shibo Wang, et~al.
\newblock Gemini 1.5: Unlocking multimodal understanding across millions of tokens of context.
\newblock {\em arXiv preprint arXiv:2403.05530}, 2024.

\bibitem{temsah2024openai}
Mohamad-Hani Temsah, Amr Jamal, Khalid Alhasan, Abdulkarim~A Temsah, and Khalid~H Malki.
\newblock Openai o1-preview vs. chatgpt in healthcare: a new frontier in medical ai reasoning.
\newblock {\em Cureus}, 16(10), 2024.

\bibitem{wang2025finnlp}
Keyi Wang, Jaisal Patel, Charlie Shen, Daniel Kim, Andy Zhu, Alex Lin, Luca Borella, Cailean Osborne, Matt White, Steve Yang, et~al.
\newblock Finnlp-fnp-llmfinlegal-2025 shared task: Regulations challenge.
\newblock In {\em Proceedings of the Joint Workshop of the 9th Financial Technology and Natural Language Processing (FinNLP), the 6th Financial Narrative Processing (FNP), and the 1st Workshop on Large Language Models for Finance and Legal (LLMFinLegal)}, pages 363--370, 2025.

\bibitem{wei2022chain}
Jason Wei, Xuezhi Wang, Dale Schuurmans, Maarten Bosma, Fei Xia, Ed~Chi, Quoc~V Le, Denny Zhou, et~al.
\newblock Chain-of-thought prompting elicits reasoning in large language models.
\newblock {\em Advances in neural information processing systems}, 35:24824--24837, 2022.

\bibitem{wu2023bloomberggpt}
Shijie Wu et~al.
\newblock Bloomberggpt: A large language model for finance.
\newblock {\em arXiv preprint arXiv:2303.17564}, 2023.

\bibitem{xie2024finben}
Qianqian Xie, Weiguang Han, Zhengyu Chen, Ruoyu Xiang, Xiao Zhang, Yueru He, Mengxi Xiao, Dong Li, Yongfu Dai, Duanyu Feng, et~al.
\newblock The finben: An holistic financial benchmark for large language models.
\newblock {\em arXiv preprint arXiv:2402.12659}, 2024.

\bibitem{xie2023pixiu}
Qianqian Xie, Weiguang Han, Xiao Zhang, Yanzhao Lai, Min Peng, Alejandro Lopez-Lira, and Jimin Huang.
\newblock Pixiu: a large language model, instruction data and evaluation benchmark for finance.
\newblock In {\em Proceedings of the 37th International Conference on Neural Information Processing Systems}, pages 33469--33484, 2023.

\bibitem{xie2024open}
Qianqian Xie, Dong Li, Mengxi Xiao, Zihao Jiang, Ruoyu Xiang, Xiao Zhang, Zhengyu Chen, Yueru He, Weiguang Han, Yuzhe Yang, et~al.
\newblock Open-finllms: Open multimodal large language models for financial applications.
\newblock {\em arXiv preprint arXiv:2408.11878}, 2024.

\bibitem{yang2024qwen2}
An~Yang, Beichen Zhang, Binyuan Hui, Bofei Gao, Bowen Yu, Chengpeng Li, Dayiheng Liu, Jianhong Tu, Jingren Zhou, Junyang Lin, et~al.
\newblock Qwen2. 5-math technical report: Toward mathematical expert model via self-improvement.
\newblock {\em arXiv preprint arXiv:2409.12122}, 2024.

\bibitem{yang2023fingpt}
Hongyang Yang, Xiao-Yang Liu, and Christina~Dan Wang.
\newblock Fingpt: Open-source financial large language models.
\newblock {\em arXiv preprint arXiv:2306.06031}, 2023.

\bibitem{yang2020finbert}
Yi~Yang et~al.
\newblock Finbert: A pre-trained financial language representation model for financial text mining.
\newblock {\em International Joint Conference on Artificial Intelligence}, 2020.

\bibitem{ye2025limo}
Yixin Ye, Zhen Huang, Yang Xiao, Ethan Chern, Shijie Xia, and Pengfei Liu.
\newblock Limo: Less is more for reasoning.
\newblock {\em arXiv preprint arXiv:2502.03387}, 2025.

\bibitem{zhang2023automatic}
Zhuosheng Zhang et~al.
\newblock Automatic chain of thought prompting in large language models.
\newblock {\em arXiv preprint arXiv:2302.12822}, 2023.

\bibitem{zhao2024docmath}
Yilun Zhao, Yitao Long, Hongjun Liu, Ryo Kamoi, Linyong Nan, Lyuhao Chen, Yixin Liu, Xiangru Tang, Rui Zhang, and Arman Cohan.
\newblock Docmath-eval: Evaluating math reasoning capabilities of llms in understanding long and specialized documents.
\newblock In {\em Proceedings of the 62nd Annual Meeting of the Association for Computational Linguistics (Volume 1: Long Papers)}, pages 16103--16120, 2024.

\bibitem{zhong2024evaluation}
Tianyang Zhong, Zhengliang Liu, Yi~Pan, Yutong Zhang, Yifan Zhou, Shizhe Liang, Zihao Wu, Yanjun Lyu, Peng Shu, Xiaowei Yu, et~al.
\newblock Evaluation of openai o1: Opportunities and challenges of agi.
\newblock {\em arXiv preprint arXiv:2409.18486}, 2024.

\bibitem{zhu2021tatqaquestionansweringbenchmark}
Fengbin Zhu, Wenqiang Lei, Youcheng Huang, Chao Wang, Shuo Zhang, Jiancheng Lv, Fuli Feng, and Tat-Seng Chua.
\newblock Tat-qa: A question answering benchmark on a hybrid of tabular and textual content in finance, 2021.

\bibitem{zhu2025dianjin}
Jie Zhu, Qian Chen, Huaixia Dou, Junhui Li, Lifan Guo, Feng Chen, and Chi Zhang.
\newblock Dianjin-r1: Evaluating and enhancing financial reasoning in large language models.
\newblock {\em arXiv preprint arXiv:2504.15716}, 2025.

\end{thebibliography}


\newpage
\appendix
\label{sec:appendix}
\section{Related Work}
Our work builds upon three main research directions: the development of general-purpose and reasoning-enhanced language models, applications of LLMs in finance, and studies on reasoning capabilities in financial tasks.
\subsection{General-Purpose and Reasoning LLMs}
Recent years have witnessed significant advancements in large language models which demonstrated remarkable capabilities across various tasks, including natural language understanding, generation, and complex reasoning. Building upon these foundations, researchers have developed specialized reasoning-enhanced models such as DeepSeek-R1 and GPT-o1, which incorporate techniques like chain-of-thought prompting \cite{wei2022chain}, self-reflection \cite{zhang2023automatic}, and reinforcement learning from human feedback \cite{ouyang2022training}. These models have shown superior performance in mathematical reasoning \cite{lewkowycz2022solving}, symbolic manipulation \cite{drori2023neural}, and logical inference tasks \cite{chen2023automatic}.
\subsection{LLMs in Financial Applications}
The application of LLMs in finance has gained significant attention, with models being adapted for tasks such as financial text analysis \cite{yang2020finbert}, market sentiment prediction \cite{yang2023fingpt}, and automated trading strategies \cite{kou2024automate}. Recent work has focused on developing finance-specific models like BloombergGPT \cite{wu2023bloomberggpt}, FinGPT \cite{yang2023fingpt} and PIXIU \cite{xie2023pixiu}, OpenFinLLMs~\cite{xie2024open}, Plutus\cite{peng2025plutus}, which are pretrained on financial corpora to better understand domain-specific terminology and concepts. These models have demonstrated effectiveness in tasks such as financial sentiment analysis, earnings call analysis, and regulatory compliance checking.

\subsection{Financial Reasoning}
Financial reasoning presents unique challenges that combine numerical computation, domain knowledge, and logical inference. Recent studies have explored LLMs' capabilities in financial reasoning tasks, including numerical reasoning over financial documents \cite{chen2021finqa}, mathematical problem-solving in financial contexts \cite{srivastava2024evaluating}, and structured data interpretation \cite{liu2025findabench}. These works highlight the importance of both domain expertise and reasoning abilities in financial applications. Our work extends these research directions by systematically evaluating how reasoning enhancements in state-of-the-art LLMs translate to financial domain tasks, providing insights into the effectiveness of different reasoning strategies and identifying areas for improvement in financial reasoning capabilities. Additionally, we focus on benchmarking these models across diverse financial datasets, emphasizing multi-table reasoning, long-context comprehension, and quantitative analysis. This comprehensive evaluation not only sheds light on current model limitations but also informs future research directions for developing more robust financial reasoning models.

\section{Details of Data Source}
\label{appendix:datasets}

\subsection{Data source examples for other works}
\label{other_data}
This section present some examples of the data from other models. In convince of reviewing, we have translated all the original context into English if they are not English. 

\begin{tcolorbox}[colback=lightgray!10, colframe=black, title=Finance-Instruct]
\fontsize{8pt}{9pt}
\begin{lstlisting}[breaklines=true, basicstyle=\ttfamily, frame=none]
Explain tradeoffs between fiscal and monetary policy as tools in a nation's economic toolkit. Provide examples of past instances when each were utilized, the economic conditions that led to them being deployed, their intended effects, and an evaluation of their relative efficacy and consequences.
\end{lstlisting}
\end{tcolorbox}

\begin{tcolorbox}[colback=lightgray!10, colframe=black, title=CFLUE]
\fontsize{8pt}{9pt}
\begin{lstlisting}[breaklines=true, basicstyle=\ttfamily, frame=none]
Assume you are a financial industry expert. Please answer the following question.
Note: The question is a single-choice question. Only return the most appropriate option. If multiple options are suitable, only return the most accurate one.
Among the following options, which financial institution is not permitted to participate in the interbank lending market?
A. Clearing Center
B. Insurance Company
C. Securities Company
D. Fund Management Company
\end{lstlisting}
\end{tcolorbox}

\begin{tcolorbox}[colback=lightgray!10, colframe=black, title=CFLUE]
\fontsize{8pt}{9pt}
\begin{lstlisting}[breaklines=true, basicstyle=\ttfamily, frame=none]
Is the sentiment of this text positive, negative, or neutral?
Feeling exhausted from averaging down.
\end{lstlisting}
\end{tcolorbox}

\begin{tcolorbox}[colback=lightgray!10, colframe=black, title=Finance-IQ]
\fontsize{8pt}{9pt}
\begin{lstlisting}[breaklines=true, basicstyle=\ttfamily, frame=none]
When considering the pricing of life insurance products, several key factors come into play, including market forces, business operating costs, and government regulations. Among these factors, which one primarily determines the minimum limit of life insurance product prices?
A. Market forces and business operating costs
B. Business operating costs
C. Government regulations
D. Market forces
\end{lstlisting}
\end{tcolorbox}

\subsection{Data source for FinCoT}

We provide here detailed descriptions of the datasets used to construct our training and evaluation corpus for financial reasoning enhancement.

\begin{itemize}
    \item \textbf{FinQA}~\cite{chen2022finqadatasetnumericalreasoning}: FinQA consists of over 5,000 open-ended question-answer pairs derived from financial reports and tables. It integrates both textual and tabular data, covering numerical reasoning tasks such as arithmetic operations, comparison, and multi-step calculations, making it well-suited for training reasoning-enhanced models.

    \item \textbf{DocFinQA}~\cite{reddy2024docfinqa}: An extension of FinQA designed for long-context reasoning. It pairs FinQA-style questions with full-length financial filings, increasing the average context length from under 700 words to approximately 123,000 words. This dataset better simulates real-world document scale and complexity, enabling robust evaluation of retrieval and reasoning over long financial documents.

    \item \textbf{TAT-QA}~\cite{zhu2021tatqaquestionansweringbenchmark}: A large-scale financial QA dataset created from real financial reports. It combines both textual and tabular data and emphasizes diverse numerical reasoning tasks, including arithmetic operations, counting, sorting, and comparison. TAT-QA reflects real-world challenges in financial document interpretation.
    
    \item \textbf{ConvFinQA}~\cite{chen2022convfinqa}: A financial question answering dataset designed to handle multi-turn conversations involving financial tables and text. ConvFinQA builds upon FinQA by incorporating follow-up questions and context retention across multiple turns, simulating more realistic financial analysis scenarios. It provides expert-annotated sub-questions that decompose complex financial questions into granular steps, facilitating the generation of structured reasoning paths and promoting systematic financial reasoning.

    \item \textbf{DocMath-Eval}~\cite{zhao2024docmath}: This benchmark suite evaluates mathematical reasoning in domain-specific contexts. We utilize two subsets:
    \begin{itemize}
        \item \textbf{DocMath-CompShort\cite{zhao2024docmath}}: Constructed from quarterly and annual reports, it involves reasoning across long documents and multiple tables, presenting a challenging multi-step numerical reasoning setting.
        \item \textbf{DocMath-SimpShort\cite{zhao2024docmath}}: Built from FinQA and TAT-QA, this subset focuses on localized numerical reasoning tasks over short documents containing a single table.
    \end{itemize}
    \item \textbf{Econ\_Logic \cite{quan2024econlogicqa}:} It consists of financial logic reasoning tasks that help the model to further understand and have more insights of financial terms and knowledge. 
    \item \textbf{BizBench-QA}~\cite{krumdick2024bizbench}: Contains financial QA pairs across multiple task types, including SEC-NUM, CodeFinQA, CodeTAT-QA, and FinCode. These tasks span code generation, numerical span identification, and financial mathematical reasoning from various perspectives, enriching the training corpus with broader financial task coverage.
\end{itemize}

\begin{tcolorbox}[colback=lightgray!10, colframe=black, title=FinQA]
\fontsize{8pt}{9pt}
\begin{lstlisting}[breaklines=true, basicstyle=\ttfamily, frame=none]
Please answer the given financial question based on the context.
Context: measurement point december 31 the priceline group nasdaq composite index s&p 500 rdg internet composite .
|measurement pointdecember 31|the priceline group inc .|nasdaqcomposite index|s&p 500index|rdg internetcomposite|
|2010|100.00|100.00|100.00|100.00|
|2011|117.06|100.53|102.11|102.11|
|2012|155.27|116.92|118.45|122.23|
|2013|290.93|166.19|156.82|199.42|
|2014|285.37|188.78|178.29|195.42|
|2015|319.10|199.95|180.75|267.25|.
Question: what was the percent of the growth of the the priceline group inc . from 2014 to 2015
Answer:
\end{lstlisting}
\end{tcolorbox}

\section{Example of CoT data curation}
\label{app:example}
Here is an example of applying the guideline and iterative refinement process to generate a CoT reasoning path based on the given context and question.

\begin{figure}
    \centering
    \includegraphics[width=0.6\linewidth]{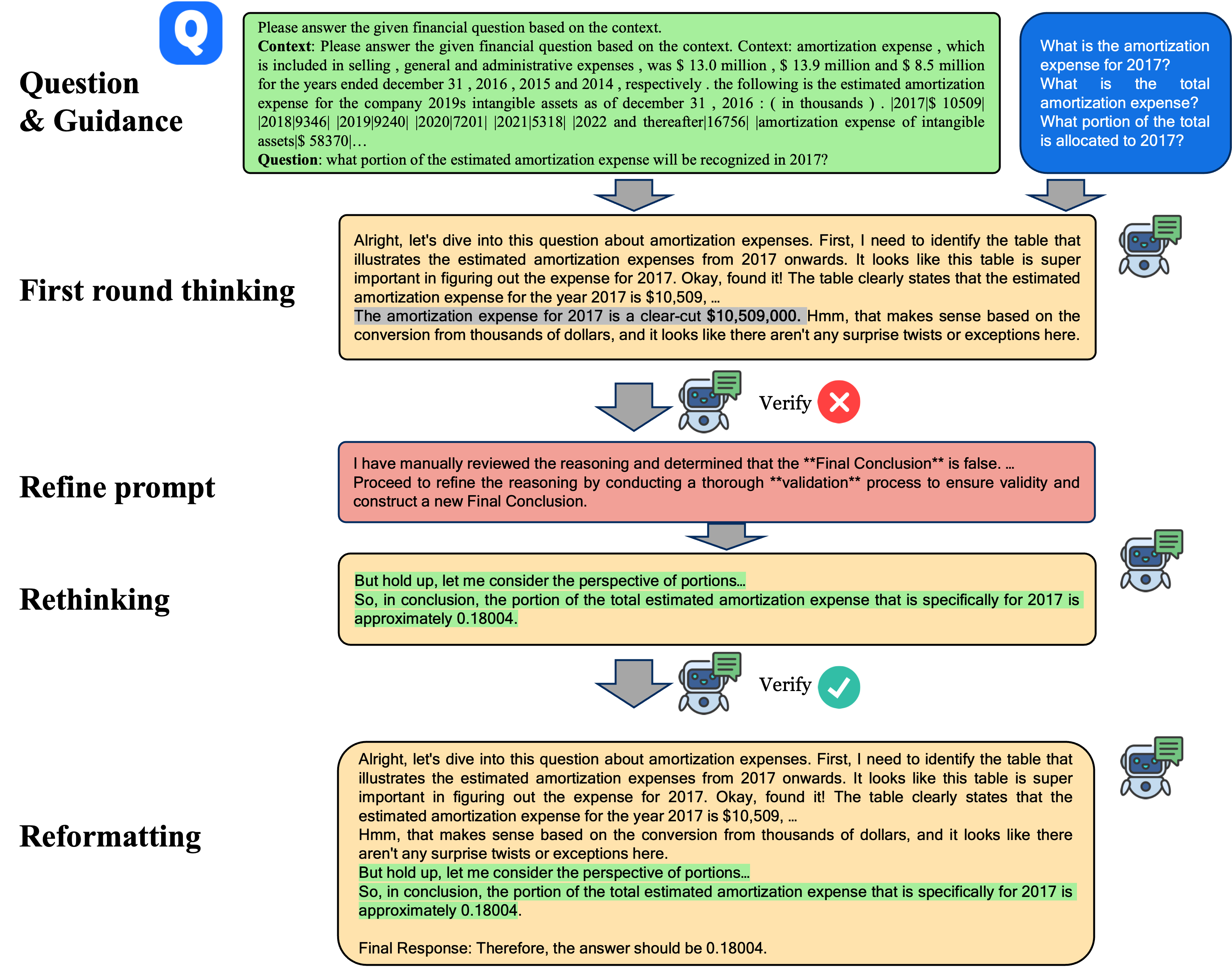}
    \caption{Example of iterative refinement of the CoT process with a guidance.}
    \label{fig:refine}
\end{figure}

\newpage
\begin{tcolorbox}[colback=lightgray!10, colframe=black, title=BizBench]
\fontsize{8pt}{9pt}
\begin{lstlisting}[breaklines=true, basicstyle=\ttfamily, frame=none]
From the results reflected in the following pro forma information because of normal production declines, changes in commodity prices, future acquisitions and divestitures, future development and exploration activities and other factors.

| | | | | | | | | | | |
| --- | --- | --- | --- | --- | --- | --- | --- | --- | --- | --- |
| | | | | | | | | | | |
| | | | | |
| (In millions, except per share amounts) | | Six Months Ended June 30, 2021 |
| Pro forma revenue | | $ | 1,802 | | | |
| Pro forma net income | | 71 | | | |
| | | | | |
| | | | | |
| | | | | |
| | | | | |
| Pro forma basic earnings per share | | $ | 0.09 | | | |
| Pro forma diluted earnings per share | | $ | 0.09 | | | |

3. Properties and Equipment, NetProperties and equipment, net are comprised of the following:

| | | | | | | | | | | | | | | |
| --- | --- | --- | --- | --- | --- | --- | --- | --- | --- | --- | --- | --- | --- | --- |
| | | | | | | | | | | | | | | |
| (In millions) | | June 30,2022 | | December 31,2021 |
| Proved oil and gas properties | | $ | 16,102 | | | $ | 15,340 | |
| Unproved oil and gas properties | | 5,292 | | | 5,316 | |
| Gathering and pipeline systems | | 423 | | | 395 | |
| Land, buildings and other equipment | | 144 | | | 140 | |
| Finance lease right-of-use asset | | 24 | | | 20 | |
| | | 21,985 | | | 21,211 | |
| Accumulated depreciation, depletion and amortization | | (4,578) | | | (3,836) | |
| | | $ | 17,407 | | | $ | 17,375 | |

As of June 30, 2022, the Company had no projects with exploratory well costs capitalized for more than one year after drilling.

| | | | | | | | | | | | | | | |
| --- | --- | --- | --- | --- | --- | --- | --- | --- | --- | --- | --- | --- | --- | --- |
| | | | | | | | | | | | | | | |
| (In millions) | | June 30,2022 | | December 31,2021 |
| 6.51% weighted-average private placement senior notes (1) | | $ | 37 | | | $ | 37 | |
| 5.58% weighted-average private placement senior notes (2) | | 87 | | | 87 | |
| 3.65% weighted-average private placement senior notes | | 825 | | | 825 | |
| 4.375% senior notes due June 1, 2024 | | 750 | | | 750 | |
| 3.90% senior notes due May 15, 2027 | | 750 | | | 750 | |
| 4.375% senior notes due March 15, 2029 | | 500 | | | 500 | |
| Revolving credit facility | | - | | | - | |
| Net premium (discount) | | 164 | | | 185 | |
| Unamortized debt issuance costs | | (8) | | | (9) | |
| | | $ | 3,105 | | | $ | 3,125 | |

(1) Includes $37 million of current portion of long-term debt at June 30, 2022.(2) Includes $87 million of current portion of long-term debt at June 30, 2022.At June 30, 2022, the Company was in compliance with all financial and other covenants for both its revolving credit facility and senior notes.
Question: What is the pro forma net income?
\end{lstlisting}
\end{tcolorbox}

\section{Details of experiments}
\subsection{Model training}
\label{detail_training}
We list the details of the training process here. 
The SFT process is configured with a learning rate of 5e-6 and a weight decay of 0.1. Training is conducted for 3 epoch with a batch size of 2 per GPU and a gradient accumulation of 2 steps, effectively resulting in an effective batch size of 2. The gradient checkpointing is enabled to optimize memory usage.
For RL,
the training process is configured with a learning rate specified set to 5e-6. The gradient accumulation is set to 2 steps, and a gradient accumulation of 2 steps, effectively resulting in an effective batch size based on the configured gradient\_accumulation\_steps. 
The training is conducted using 3 GPUs, each handling a separate training process. And a total of 500 steps is used to train the model. 
\subsection{Model evaluation}
\label{model_evaluation}
For large models (e.g., DeepSeek-V3, DeepSeek-R1, GPT series, Llama4), we run evaluations via APIs from TogetherAI\footnote{\url{https://www.together.ai}} and OpenAI\footnote{\url{https://platform.openai.com}}, using versions GPT-o1-2024-12-17, GPT-o3-mini-2025-01-31, and GPT-4o-2024-08-06. All open-sourced models up to 72B parameters were conducted with vLLM \cite{kwon2023efficient} framework on 8 A100 GPUs (80GB each), with a max generation length of 1,024 tokens and temperature setting of 0. Given the extreme length of some evaluated dataset (e.g., Dm-Complong with a average length of 40K tokens), we use a \texttt{zero-shot} setting, aim to evaluating LLMs' capabilities for financial reasoning without any few-shot demonstrations.

\section{Details of Evaluated Models}
\label{model_description}

To comprehensively evaluate the performance of LLMs in financial tasks, particularly those with reasoning enhancements, we select a diverse set of 29 LLMs. This selection includes both general-purpose, and finance-specific models, ranging from smaller models (8B) to large-scale models (70B), and extending to ultra-large models such as {DeepSeek-R1} (671B), {GPT-4o}, and {GPT-o1}. 
The evaluated models are summarized as follows:
\textbf{GPT-4o}~\cite{hurst2024gpt}: OpenAI’s flagship model with multimodal capabilities and improved real-time reasoning.
\textbf{GPT-o1}~\cite{jaech2024openai}: A reasoning-optimized variant in the GPT series with strong step-by-step problem-solving ability.
\textbf{GPT-o3-mini}: A compact version of GPT-o3, optimized for efficiency with robust language understanding.
\textbf{DeepSeek-V3}~\cite{liu2024deepseek}: A 671B Mixture-of-Experts (MoE) model with 37B active parameters per token, designed for efficient reasoning.
\textbf{DeepSeek-R1}~\cite{guo2025deepseek}: A first-generation reasoning model with competitive performance on math, code, and reasoning tasks.
\textbf{Qwen series}~\cite{yang2024qwen2}: A strong multilingual model with advanced instruction tuning.
\textbf{LLaMA-series}~\cite{dubey2024llama}: Widely adopted open-source models (8B–70B) as well as the latest MOE models with strong instruction-tuned variants used in our experiments.
\textbf{DeepSeek-R1-Distill}~\cite{guo2025deepseek}: Distilled models from DeepSeek-R1, fine-tuned using LLaMA and Qwen backbones to balance performance and efficiency.
\textbf{Financial-Specific Reasoning Models}: This category includes models such as \textbf{Fin-R1-7B} \cite{finr1} and \textbf{Dianjin-R1-7B} \cite{zhu2025dianjin}, both of which leverage financial reasoning datasets and employ fine-tuning and reinforcement learning techniques to optimize LLM performance in financial reasoning tasks.
\textbf{Other reasoning models}: They include {S1-32B} \cite{muennighoff2025s1}, {Limo}~\cite{ye2025limo}. These two models are designed for general reasoning tasks, with limited but high quality COT data.

More details of the evaluated models are presented in Table~\ref{tab:LLMs}. It also highlights the diverse training strategies employed across the evaluated models. Reinforcement learning has been extensively utilized to enhance reasoning capabilities, with common methods including PPO, DPO, and GRPO. Among these, GRPO is the most widely adopted, as seen in models such as DeepSeek-R1, Qwen2.5-72B-Instruct, and FinR1-7B. This widespread use of GRPO underscores its effectiveness in refining reasoning abilities, particularly in complex, multi-step tasks involving financial documents.

\begin{table}[t]
    \centering
    \caption{Summary of evaluated LLMs, including their parameter sizes, reasoning capabilities, input limits, source availability, and reasoning enhancement strategies. MCTS refers to Monte Carlo Tree Search. CoT refers to chain-of-thought reasoning. TIR means tool-integrated reasoning.}
    \label{tab:LLMs}
    \scriptsize
    \setlength{\tabcolsep}{3pt} 
    \renewcommand{\arraystretch}{1} 
    \resizebox{0.9\columnwidth}{!}{%
    \begin{tabular}{>{\raggedright}p{3.3cm} >{\centering}p{1.1cm} >{\centering}p{2.1cm} >{\centering}p{1.1cm} >{\centering}p{0.9cm} >{\centering\arraybackslash}p{3.1cm}}
        \hline
        \textbf{Model Name} & \textbf{Parameters} & \textbf{Training} & \textbf{Window Size} & \textbf{Close/Open Source} & \textbf{Reasoning Training Data} \\
        \hline
        GPT-4o & - & - & 128k & Closed & - \\
        GPT-o1 & - & RL with long CoT & 128k & Closed & Public and proprietary data (Human-annotated CoT, MCTS-assisted Synthetic data)  \\ 
        GPT-o3-mini & - & - & 128k & Closed & Public and proprietary data \\
        DeepSeek-V3 & 671B &  GRPO & 128k & Open & -  \\
        DeepSeek-R1 & 671B & GRPO & 128k & Open & Cold-start data generation, post-processing data \\
        GPT-4.5 & - & - & 128k & Closed & - \\
        \hline
        Llama4-Scout & 17B*16E & SFT \& DPO & 128k & Open & -  \\
        Qwen2.5-72B-Instruct & 72B & DPO \& GRPO & 128k & Open & -  \\
        Qwen2.5-72B-Instruct-Math & 72B & DPO \& GRPO & 128k & Open & Synthetic data from Qwen, high-quality mathematical data, CoT, TIR \\
        DeepSeek-R1-Distill-Llama-70B & 70B & GRPO & 128k & Open & Distilled from R1 \\
        Llama3-70B-Instruct & 70B &  SFT \& DPO & 8k & Open & - \\
        Llama3.1-70B-Instruct & 70B &  SFT \& DPO & 128k & Open & - \\
        Llama3.3-70B-Instruct & 70B &  SFT \& DPO & 128k & Open & - \\
        \hline
        Qwen2.5-32B-Instruct & 32B & DPO \& GRPO & 128k & Open & - \\
        Qwen3-32B & 32B & RL with long CoT & 128k & Open & - \\
        DeepSeek-R1-Distill-Qwen-32B & 32B & GRPO & 128k & Open & Distilled from R1 \\
        Limo & 32B & SFT & 128k & Open & Public data with Qwen filteration, with CoT generated from Deepseek \\
        S1-32B & 32B & SFT & 128k & Open & Public data and original data, filtered by quality, with CoT generated from Gemini \\
        Qwen/QwQ-32B & 32B & PPO & 128k  & Open & - \\
        \hline
        Qwen2.5-14B-Instruct & 14B & DPO \& GRPO & 128k & Open & - \\
        Qwen3-14B & 14B & RL with long CoT & 128k & Open & - \\
        DeepSeek-R1-Distill-Qwen-14B & 14B & GRPO & 128k & Open & Distilled from R1 \\
        DeepSeek-R1-Distill-Llama-8B & 8B & GRPO & 128k & Open & Distilled from R1 \\
        Llama3-8B-Instruct & 8B & SFT \& DPO & 8k & Open & - \\
        Llama3.1-8B-Instruct & 8B & SFT \& DPO & 128k & Open & - \\
        Qwen2.5-7B-Instruct & 7B & DPO \& GRPO & 128k & Open & - \\
        Qwen3-8B & 8B & - & 128k & Open & - \\
        FinR1-7B & 7B & SFT \&  GRPO & 128k & Open & 60k CoT data for financial tasks \\
        Dianjin-R1-7B & 7B & SFT \& GRPO & 128k & Open & 31k CoT data for financial tasks \\ 
        \hline
    \end{tabular}
    }
\end{table}


\begin{figure}[htbp]
    \centering
    \includegraphics[width=0.5\textwidth]{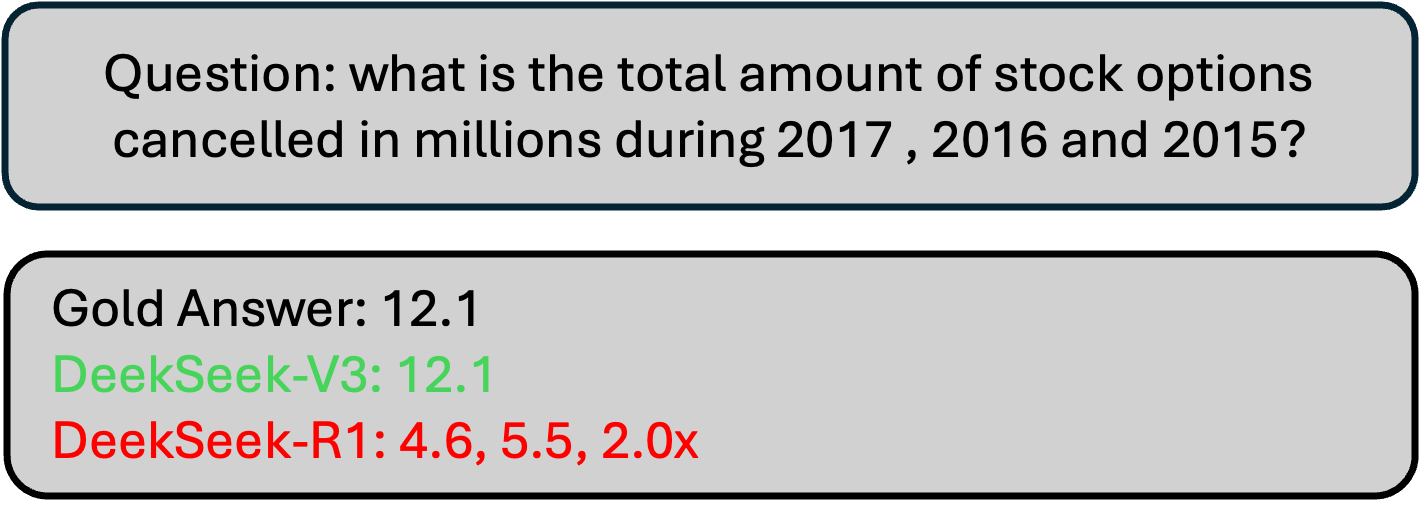}
    \caption{Error case 1.}
    \label{fig:case1}
\end{figure}

\begin{figure*}
    \centering
    \includegraphics[width=1\textwidth]{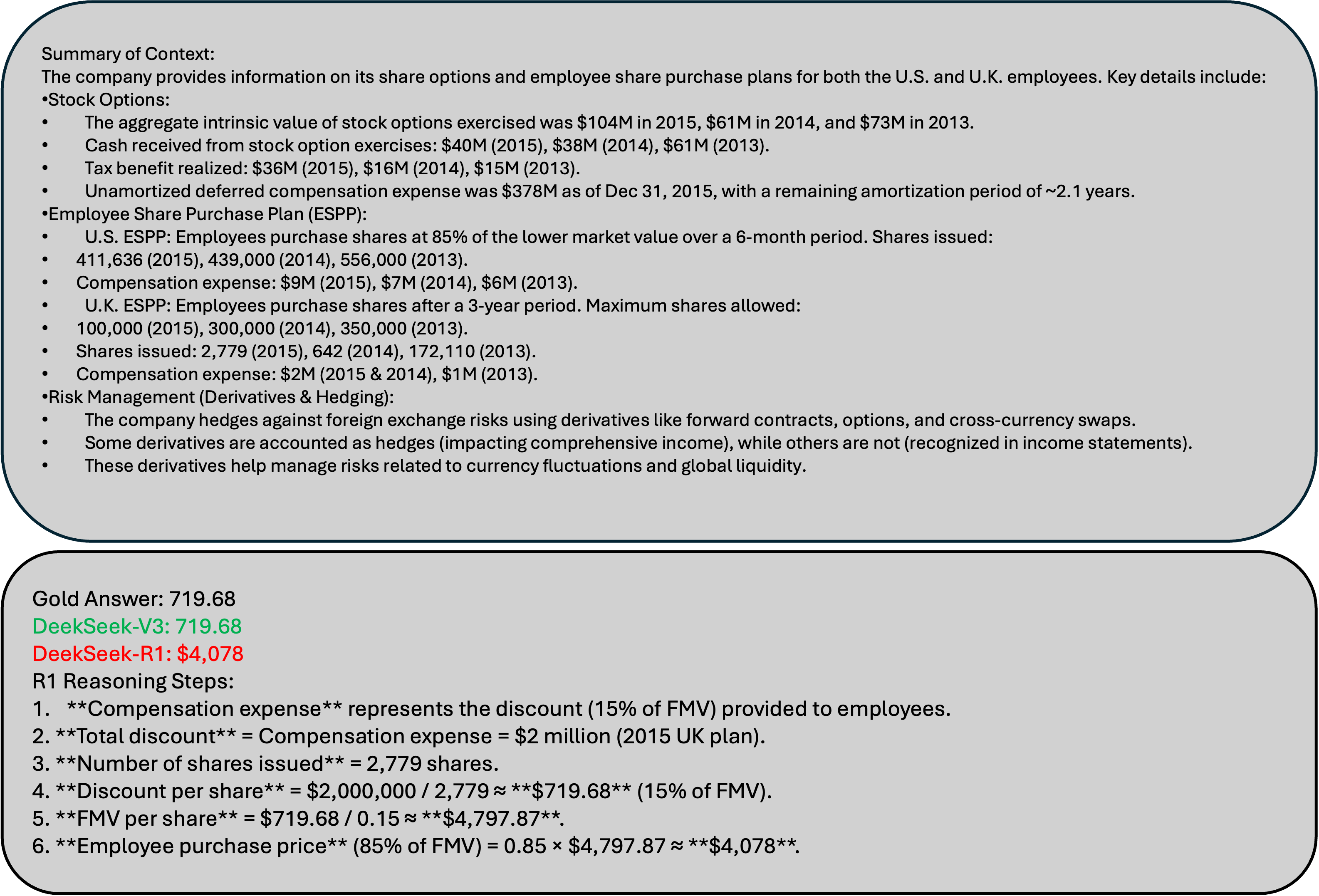}
    \caption{Error case 2.}
    \label{fig:case2}
\end{figure*}

\section{Error Analysis}
To further investigate why reasoning-enhanced models like DeepSeek-R1 and GPT-o1 underperform in financial tasks compared to general-purpose models such as GPT-4o and DeepSeek-V3, we conduct an error analysis using the FinQA dataset since these reasoning-enhanced models exhibit a significant performance drop on FinQA.

We focus on cases where DeepSeek-R1 produces incorrect answers while DeepSeek-V3 generates correct ones. Two common error patterns emerge. First, DeepSeek-R1 tends to over-reason and does not strictly adhere to instructions, as illustrated in Figure \ref{fig:case1}. In this case, the question requires a sum, but the model provides excessive details by breaking down the calculation instead of directly summing the values as instructed.

Second, compared to DeepSeek-V3, DeepSeek-R1 appears to lack financial sensitivity. For instance, in Figure \ref{fig:case2}, it fails to recognize critical financial nuances, leading to an incorrect response. Specifically, DeepSeek-R1 misinterprets the concept of "average share price" in the UK Employee Share Purchase Plan by incorrectly identifying it as the employee purchase price (\$4,078 per share) rather than the actual compensation expense per share (\$719 per share). This mistake stems from an inability to correctly differentiate between compensation expense per share and the full fair market value (FMV) calculation, resulting in an overestimated figure.
In contrast, DeepSeek-V3 correctly understands that the compensation expense per share represents the direct financial impact per issued share, allowing it to provide the accurate answer. This highlights a fundamental gap in DeepSeek-R1’s financial reasoning capabilities, particularly in recognizing the accounting conventions used to report share-based compensation.

\section{Prompt}
\label{app:prompt}
\begin{tcolorbox}[colback=lightgray!10, colframe=black, title=Prompt Template for DocMath-CompLong task]
\fontsize{8pt}{9pt}
\begin{lstlisting}[breaklines=true, basicstyle=\ttfamily, frame=none]
You are a financial expert, you are supposed to answer the given question based on the provided financial document context. You need to first think through the problem step by step, documenting each necessary step. Then you are required to conclude your response with the final answer in your last sentence as 'Therefore, the answer is {final answer}'. The final answer should be a numeric value.
###Context
<Full document here>

### Input
<the question based on the document>

Let's think step by step to answer the given question.

### Output
\end{lstlisting}
\end{tcolorbox}

\begin{tcolorbox}[colback=lightgray!10, colframe=black, title=Prompt Template for  DocMath-Simplong task]
\fontsize{8pt}{9pt}
\begin{lstlisting}[breaklines=true, basicstyle=\ttfamily, frame=none]
You are a financial expert, you are supposed to answer the given question based on the provided financial document context. You need to first think through the problem step by step, documenting each necessary step. Then you are required to conclude your response with the final answer in your last sentence as 'Therefore, the answer is {final answer}'. The final answer should be a numeric value.

###Context
<simple long context here>

### Input
<the question based on the context>

Let's think step by step to answer the given question.

### Output
\end{lstlisting}
\end{tcolorbox}

\begin{tcolorbox}[colback=lightgray!10, colframe=black, title=Prompt Template for FinQA task]
\fontsize{8pt}{9pt}
\begin{lstlisting}[breaklines=true, basicstyle=\ttfamily, frame=none]
Please answer the given financial question based on the context.
Context: <context>
Question: <the question based on the context>
Answer:
\end{lstlisting}
\end{tcolorbox}

\begin{tcolorbox}[colback=lightgray!10, colframe=black, title=Prompt Template for XBRL-Math task]
\fontsize{8pt}{9pt}
\begin{lstlisting}[breaklines=true, basicstyle=\ttfamily, frame=none]
You are a financial expert tasked with carefully reading, analyzing, and answering the following eXtensible Business Reporting Language. Please follow the steps below:

INPUT: Read the eXtensible Business Reporting Language (XBRL) question: <XBRL related question>, formula: <the formula related to this question>, and the explanation: <term explanation>. Provide only the final answer which is the numerical result of the calculation. For formulas like ROI, provide percentages. Never use the percent symbol in percentages.

OUTPUT:
\end{lstlisting}
\end{tcolorbox}



\section*{Limitations}
\label{limitation}
Our study has several limitations that highlight areas for future improvement. First, our model scale is limited, as we only fine-tuned an 8B model (Fino1), while larger models (e.g., 70B) could potentially benefit more from reasoning enhancements. Second, our evaluation scope is restricted, covering only three financial reasoning tasks (FinQA, DM-Simplong, and XBRL-Math), which do not fully capture the breadth of financial NLP applications such as forecasting, financial sentiment analysis, and fraud detection. Third, our fine-tuning relies on a single dataset, FinQA, for reasoning path construction, limiting the model’s exposure to different financial reasoning patterns; incorporating additional datasets could improve generalization. Fourth, our reasoning path construction approach is simplified, as we generate paths using a single method (GPT-4o), whereas exploring multiple reasoning path generation strategies—such as ensemble approaches or human-annotated paths—could lead to more robust financial reasoning capabilities. 

\section*{Future Direction}
Developing effective reasoning models for financial tasks poses distinct challenges compared to general-domain reasoning. Our findings suggest three key directions for improvement: (1) Enhancing financial knowledge and terminology understanding through domain-specific pretraining using financial corpora, structured reports, and regulatory documents; (2) Improving multi-table reasoning and long-context comprehension by developing strategies that better infer logic across tables and optimize memory mechanisms for processing lengthy financial documents; and (3) Refining reasoning-enhanced strategies, as current approaches like GPT-o1’s chain-of-thought prompting and DeepSeek-R1’s self-reflection via reinforcement learning yield mixed results—highlighting the need for more tailored reasoning strategies to address both structured numerical and unstructured textual financial reasoning.

\section*{Ethics Statement}

While our study demonstrates significant advancements in financial reasoning with LLMs, Fino1 remains a research prototype and is not yet suitable for real-world financial applications. Our model inherits the well-documented limitations of large language models, including hallucinations, sensitivity to input phrasing, and potential biases in financial data sources, which could affect reliability in critical financial contexts. Furthermore, Fino1 has not undergone rigorous testing for high-stakes financial decision-making, such as investment strategies, regulatory compliance, or risk assessment, where incorrect outputs could lead to severe consequences. Additionally, the model’s reliance on a single dataset (FinQA) and a single reasoning path construction method (GPT-4o) limits its robustness and adaptability to the complex and evolving nature of financial environments. As with all AI-driven financial models, careful human oversight remains essential, and future work must focus on enhancing factual consistency, mitigating biases, improving domain adaptation, and ensuring alignment with regulatory frameworks before such models can be considered for real-world financial deployment.

\section*{Potential impacts}
\label{impact}
The proposed work presents both positive and negative societal implications. The development of advanced financial reasoning models has the potential to enhance financial analysis by providing more accurate predictions and robust risk assessments, thereby supporting informed decision-making in areas such as investment strategies and regulatory compliance. However, these models may also pose risks if misused, such as reinforcing existing biases in financial datasets or facilitating manipulative financial practices. To mitigate such risks, we incorporate safeguards that enforce controlled access to the models, implement guidelines for responsible use, and maintain transparency in model outputs. Additionally, we emphasize the importance of adhering to ethical guidelines throughout the development and deployment process to minimize unintended societal consequences.

\newpage

\end{document}